\RequirePackage{amsmath}
\documentclass[runningheads]{llncs}

\usepackage[T1]{fontenc}

\usepackage{graphicx}

\usepackage{bbding}

\usepackage{hyperref}
\usepackage{color}

\usepackage{amssymb}
\usepackage{amsmath}
\usepackage{bm}

\usepackage{booktabs}
\usepackage{multirow}
\usepackage{tabularx}
\usepackage{makecell}
\newcolumntype{Y}{>{\centering\arraybackslash}X}
\usepackage{threeparttable}
\usepackage{tablefootnote}

\usepackage{pifont}

\usepackage[inline]{enumitem}

\begin{document}

\title{\texorpdfstring{$n$}~-gram injection into transformers for~dynamic~language model adaptation in~handwritten~text~recognition}
\titlerunning{\texorpdfstring{$n$}~-gram injection into transformers for dynamic LM adaptation}

\author{Florent Meyer\inst{1,2}~\Envelope~~\orcidID{0009-0001-7527-5215} \and
Laurent Guichard\inst{1}\orcidID{0009-0008-7853-8704} \and
Yann Soullard\inst{2}\orcidID{0009-0001-8048-2489} \and
Denis Coquenet\inst{2}\orcidID{0000-0001-5203-9423} \and
Guillaume Gravier\inst{2}\orcidID{0000-0002-2266-5682} \and
Bertrand Coüasnon\inst{2}\orcidID{0000-0002-7077-0751}}
\authorrunning{F. Meyer et al.}
\institute{ANTAI, Rennes, France \and
Univ Rennes, CNRS, IRISA - UMR 6074, Rennes, France 
\\
\email{florent.meyer@irisa.fr}}

\maketitle

\begin{abstract}
Transformer-based encoder-decoder networks have recently achieved impressive results in handwritten text recognition, partly thanks to their auto-regressive decoder which implicitly learns a language model. However, such networks suffer from a large performance drop when evaluated on a target corpus whose language distribution is shifted from the source text seen during training. To retain recognition accuracy despite this language shift, we propose an external $n$-gram injection (NGI) for dynamic adaptation of the network's language modeling at inference time. Our method allows switching to an $n$-gram language model estimated on a corpus close to the target distribution, therefore mitigating bias without any extra training on target image-text pairs. We opt for an early injection of the $n$-gram into the transformer decoder so that the network learns to fully leverage text-only data at the low additional cost of $n$-gram inference. Experiments on three handwritten datasets demonstrate that the proposed NGI significantly reduces the performance gap between source and target corpora.

\keywords{Language model \and Transformer \and Knowledge injection \and Linguistic bias \and $n$-gram \and Handwritten text recognition}
\end{abstract}

\setcounter{footnote}{0} 

\section{Introduction}

In an era of digitalization, handwritten text still plays a crucial role in our daily lives as not all tasks and procedures have been dematerialized. For instance, a variety of handwritten content can be found in historical records, administrative forms, personal documents or complaints addressed to companies. Handwritten text recognition (HTR) is the computer vision discipline aiming at transcribing scans of these handwritten documents to machine-readable text. The recognition is still challenging as of today because of variations in writers' style, ambiguities arising from cursive scripts, or syntactic and grammar rules that differ from one language to another. 
Nowadays, transformer-based~\cite{vaswaniAttentionAllYou2017} encoder-decoders~\cite{barrereLightTransformerBasedArchitecture2022,coquenetDANSegmentationfreeDocument2023,liTrOCRTransformerbasedOptical2023,michaelEvaluatingSequencetoSequenceModels2019} achieve remarkably low transcription error rates in HTR thanks to an auto-regressive modeling of the language at training time.

When designing HTR systems, it is most often assumed that the training and test datasets follow distributions which are independent and identical~\cite{garrido-munozGeneralizationHandwrittenText2025}. However, real-world use cases do not always satisfy this constraint. Consequently, overlooking the language \textit{shift} between corpora can lead to a significant recognition accuracy drop~\cite{barrereTrainingTransformerArchitectures,kangCandidateFusionIntegrating2021,pippiHowChoosePretrained2023}, making the recognizer unusable on an out-of-distribution corpus. Our experiments on state-of-the-art encoder-decoders confirm the performance degradation of transformers on a test corpus whose language distribution differs from training.

In this paper, we propose $n$-gram injection (NGI) as a domain adaptation method to reduce the performance gap between a training dataset, referred to as \textit{source}, and a \textit{target} dataset, i.e. that seen at test time, whose language distribution is shifted. NGI is a lightweight method for guiding a neural recognizer with an external $n$-gram language model (LM). By injecting $n$-gram distributions into a new lightweight auto-regressive FCN-transformer, the word attention network (WAN), we let the network learn to adapt to the external LM. Most importantly, our method allows the user to dynamically switch the $n$-gram at test time to better match the shifted language distribution of a target corpus. Therefore, bias and the associated performance loss are mitigated without any extra training on target images.
 
To match the shift naturally present in an industrial form recognition task, we provide our own splits of well-known HTR datasets where lexicons or characters sequences differ between source and target corpora. Extensive evaluation reports large gains on target subsets while preserving performance on source subsets, demonstrating that NGI alleviates bias towards the source language. While evaluated on word-level HTR, the proposed $n$-gram injection framework could be applied to any architecture relying on an auto-regressive character-level transformer decoder.

\section{Related work}

\subsection{Bias in handwritten text recognition}

Although remarkably low error rates are obtained on printed text nowadays, HTR remains a challenging research problem as handwriting is arguably more diverse and complex than printed text. Early on, recurrent neural networks (RNN) and more particularly long short-term memory (LSTM), trained thanks to the connectionist temporal classification loss~\cite{gravesConnectionistTemporalClassification2006}, have been widely adopted by the community of document analysis. The bidirectional LSTM~\cite{gravesNovelConnectionistSystem2009,soullardImprovingTextRecognition2019} and multidimensional LSTM~\cite{blucheA2iAArabicHandwritten2014,gravesOfflineHandwritingRecognition2008} are notable examples of such recurrent architectures. Fully convolutional networks (FCN) then appeared as a means to overcome the latency caused by the sequential nature of RNN~\cite{coquenetRecurrencefreeUnconstrainedHandwritten2020,yousefFCN2020}.
Departing from recurrent and convolutional architectures, transformers~\cite{vaswaniAttentionAllYou2017} equipped with attention mechanisms implicitly build language modeling thanks to their auto-regressive decoding process, therefore achieving remarkable performance in HTR~\cite{barrereLightTransformerBasedArchitecture2022,liTrOCRTransformerbasedOptical2023,michaelEvaluatingSequencetoSequenceModels2019}.
Following this trend, the DAN~\cite{coquenetDANSegmentationfreeDocument2023}, combining a CNN encoder with a transformer decoder, has become the reference for HTR on full-page scans.

Despite their high performance, those neural networks (NN) model source language seen during training and thus suffer from a performance gap when evaluated on a target corpus whose language is shifted from the source, as shown by diverse work in scene text recognition~\cite{huVisionLanguageAdaptiveMutual2023,parkImprovingSceneText2023,wanVocabularyRelianceScene2020a,zhangContextBasedContrastiveLearning2022,zhengMRNMultiplexedRouting2023} and HTR~\cite{barrereTrainingTransformerArchitectures,kangCandidateFusionIntegrating2021,pippiHowChoosePretrained2023}. 
Yet, Garrido-Munoz \& Calvo-Zaragoza~\cite{garrido-munozGeneralizationHandwrittenText2025} relate that current research on network architectures is mostly oriented towards increasing performance on standard benchmarks, making the strong hypothesis of an independent and identical distribution between source and target corpora. However, this is overoptimistic in real-world use cases. They demonstrated that the difference between source and target character error rates reaches an astounding 37.6 points averaged on seven widely-used HTR datasets.

\subsection{Combining language models with neural networks}

\subsubsection{Language model variants}

LM can be categorized into two families, namely statistical and NN LM. 
On the one hand, $n$-grams are probabilistic models assuming that the next token in a sequence depends on a fixed-sized window of the $n-1$ previous tokens~\cite{jurafskySpeechLanguageProcessing2014}. Estimated using token counts, they provide a probability distribution over all possible tokens, making use of smoothing algorithms to avoid giving a zero probability to unseen sequences. Used in various natural language processing (NLP) tasks for several decades, $n$-grams have been the reference for the post-processing of the optical recognition in HTR~\cite{soullardImprovingTextRecognition2019,tarrideRevisitingNGramModels2024,voigtlaenderHandwritingRecognitionLarge2016}.
On the other hand, external NN LM have the key ability to capture long-range relationships between tokens. However, the training and inference of a neural LM comes at a much higher cost due to its number of parameters, while $n$-grams suffice to achieve satisfactory performance on diverse use cases. Hence, they still are a popular choice in recent work~\cite{liuInfinigramScalingUnbounded2024,tarrideRevisitingNGramModels2024}.

\subsubsection{Implicit language modeling}

Numerous forms of knowledge injection were proposed to provide NN with information about the structure of the input tokens sequence. To better model the strong correlations between adjacent elements in music sequences, Tian \textit{et al.}~\cite{tianNgramUnsupervisedCompoundation2024} group music-specific tokens before embedding them and sending them to a masked-language-modeling transformer encoder. In a similar manner, text tokens can be grouped based on $2$-grams~\cite{royNGrammerAugmentingTransformers2022} or a word lexicon~\cite{diaoZENPretrainingChinese2020,diaoTamingPretrainedLanguage2021} to encourage the network to use these related tokens together. Conversely, some work explores the reinforcement of the implicit language modeling of an NN by predicting groups of tokens~\cite{Coquenet2026,tassopoulouEnhancingHandwrittenText2021}. Similarly in pure NLP, an $n$-gram masking method enhances the integration of language information into masked language modeling by combining three prediction objectives, namely predicting $n$-gram identities as well as each separate token composing the $n$-gram, and detecting tokens replaced by a generator~\cite{xiaoERNIEGramPreTrainingExplicitly2021}. Such models require vast amounts of data and training time since they are entirely NN-based, and HTR approaches would require annotated text-image pairs to counteract language shift.

\subsubsection{Explicit language modeling}

\paragraph{Pre-processing}A family of NLP methods proposes to accelerate the costly auto-regressive inference step of transformer decoders by refining a preliminary draft issued by a faster model. The draft, which can be seen as a pre-processing step, is a tokens sequence suggested either by a standard $n$-gram~\cite{leviathanFastInferenceTransformers2023} or an LLM~\cite{stewartNGrammysAcceleratingAutoregressive2024}. The main model, a larger transformer, then verifies and corrects the suggestions, delivering speed gains whenever low-cost draft tokens are deemed valid. As opposed to sampling discrete tokens from the preliminary model, we argue that keeping the empirical distribution over all tokens provides the main network with richer information.

\paragraph{Post-processing} Numerous approaches post-process the NN output lattice \textit{via} re-scoring with the Viterbi algorithm~\cite{soullardImprovingTextRecognition2019} or beam search~\cite{swailehHandwritingRecognitionMultigrams2017,tarrideRevisitingNGramModels2024}
on characters, sub-lexical units or words. Notably, re-scoring with $n$-grams was shown to improve HTR performance of a transformer model~\cite{tarrideRevisitingNGramModels2024}. Unfortunately, re-scoring computations explore large search graphs or lattices which induces a time complexity overhead. Also, post-processing does not let the NN learn to make wise use of the LM, which leads us to think there may exist better ways of combining the two. Indeed, post-processing is applied with fixed hyper-parameters regardless of the input image, while we suppose that an NN could adjust the integration of the LM depending on visual ambiguities or in the presence of a deteriorated scan, for instance. 

\paragraph{Layer injection}Some approaches propose the injection of external LM scores to be processed by an NN model. Examples include external LM based on RNN~\cite{gulcehreUsingMonolingualCorpora2015,kangCandidateFusionIntegrating2021} which, as mentioned previously, require large amounts of training data and computing power. Furthermore, late fusion~\cite{gulcehreUsingMonolingualCorpora2015} of the LM does not allow the NN to fully exploit it, and the vision-language co-adaptation proposed by Kang \textit{et al.}~\cite{kangCandidateFusionIntegrating2021} cannot be learnt on a target domain without text-image pairs. As opposed to Li \textit{et al.}~\cite{liResidualLearningNeural2022}, our goal is not to concentrate the network's language modeling effort on difficult cases and long-distance dependencies, but precisely to restrict implicit language modeling which would restrain language adaptation capabilities. Moreover, for each different domain, the same $n$-gram is seen during both training and testing, while our objectives is to allow switching to an \textit{unseen} $n$-gram. The integration of  $n$-grams within transformers remains mostly unexplored~\cite{tarrideRevisitingNGramModels2024}, arguably due to the supposedly satisfying language modeling of the auto-regressive architecture. In this work, we try to fill this gap by introducing the first language adaptation strategy for HTR relying on $n$-gram injection.

\section{Context \& problem statement}

Recently, encoder-decoder architectures based on transformers~\cite{vaswaniAttentionAllYou2017} have achieved remarkable performance in HTR~\cite{barrereLightTransformerBasedArchitecture2022,coquenetDANSegmentationfreeDocument2023,liTrOCRTransformerbasedOptical2023,michaelEvaluatingSequencetoSequenceModels2019}. However, as shown later on and already demonstrated in the literature~\cite{barrereTrainingTransformerArchitectures}, they are prone to bias which causes a significant drop in performance between a source and a target dataset when those datasets have different language distributions. This bias can be expressed by an increase in perplexity, that is: 
\begin{equation}
\text{PPL}_{\text{LM}}(\mathcal{T}) \gg \text{PPL}_{\text{LM}}(\mathcal{S})
\end{equation}
where $\text{PPL}_{\text{LM}}(\mathcal{T})$ and $\text{PPL}_{\text{LM}}(\mathcal{S})$ are character-level perplexities of an external LM estimated on a source dataset $\mathcal{S}$ and evaluated on a target $\mathcal{T}$ or a source dataset $\mathcal{S}$, respectively. Such a language shift can appear regardless of the visual uniformity between the images in the source and target datasets, for instance when the same writer is asked to fill several fields on a personal data form. In this case, the writing can be visually very similar while the language varies a lot between name, surname and locality, for instance. In this regard, we henceforth name \textit{subsets} two sets of images which are visually similar as they originate from the same dataset but differ in their language distributions.

We argue that by learning to model the language of the source subset corpus at training time, an NN becomes biased towards it. While this bias can provide satisfying performance on a source test set, it prevents decent recognition accuracy on a target test set whose language is shifted. Yet, we hypothesize that character features learnt by a vision-encoder are robust enough even if the encoder is trained on source images only. Hence, we focus on the decoder which learns to model language \textit{via} auto-regression. Our goal is to reduce the influence of its implicit modeling of source language so that it instead relies on an external LM which we can exchange at inference time. We aim for a system which correctly reads scans from a shifted target subset \textit{without training it on target image-text pairs}, instead leveraging \textit{unpaired} text data.

\section{$n$-gram injection for dynamic language model adaptation}

In the present section, we first introduce our $n$-gram injection framework as a means to alleviate the performance drop on target data. This framework can complement any character-level encoder-decoder whose decoder is transformer-based. Second, we present the lightweight transformer-based encoder-decoder into which we inject the $n$-gram.

\subsection{$n$-gram injection}

\begin{figure}
\centering
\includegraphics[width=1\textwidth]{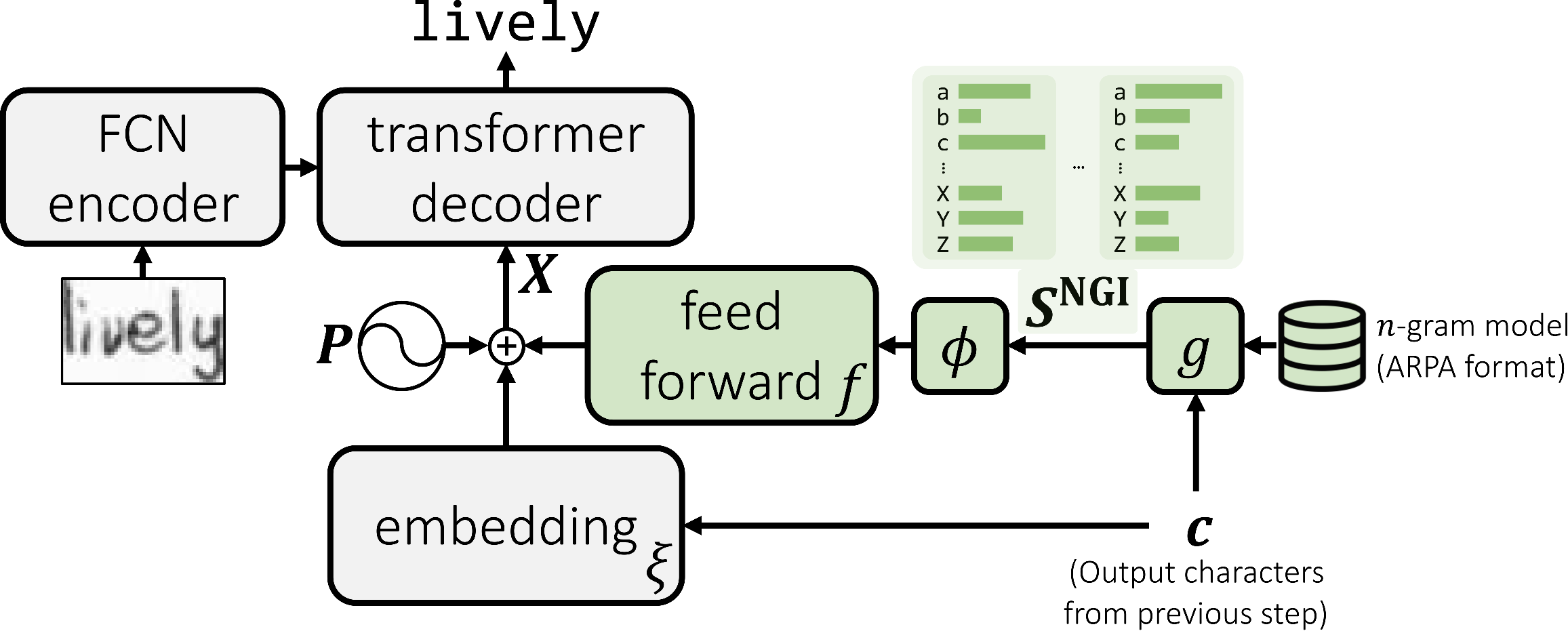}
\caption{\textbf{$n$-gram injection (NGI) into the auto-regressive decoder of the word attention network (WAN).} An $n$-gram distribution vector $\bm{S}^\text{NGI}$ is made noisy by function $\phi$, projected by $f$ then summed with standard embeddings of previously output characters $\xi(\bm{c})$ and positional encoding $\bm{P}$ to form the new decoder input $\bm{X}$. Switching to an appropriate $n$-gram at test time reduces the error gap on target corpora.} \label{fig:archi}
\end{figure}

\subsubsection{Method}

Let us now introduce our early injection mechanism of a character-level $n$-gram into an auto-regressive transformer decoder, as illustrated in Fig.~\ref{fig:archi}. Such decoders commonly predict the next token in a sequence based on the embeddings and positional encoding of previously predicted tokens. We complement this standard framework by adding language information from an $n$-gram into the input of the decoder.

Starting from log-probabilities and back-off weights stored in ARPA format, $n$-gram scores of all characters given a context are computed then inverse-log transformed and grouped into an $n$-gram distribution vector denoted by $\bm{s}^\text{NGI}_t$. At decoding step $t$, this is expressed by the following equation:
\begin{equation}
    \begin{gathered}
    s_{t,k} = g([c_{t-(n-1)}, \ldots, c_{t-1}]^{\top}) = \mathbb{P}(c_{t,k}|c_{t-(n-1)}, \ldots, c_{t-1}) \\
    \bm{s}^\text{NGI}_t = [s_{t,1}, \ldots, s_{t,K} ]^{\top} 
    \end{gathered}
\end{equation}
where $c_{t,k}$ is a character relating to position $t$ in the sequence with $k$ the index of a character among character set $K$, and $g$ is a function which maps a context vector $[c_{t-(n-1)}, \ldots, c_{t-1}]^{\top}$ to $n$-gram probabilities for each character based on log-probabilities and back-off weights read in the ARPA file, assuming that $t-(n-1) \geq 0$, otherwise the context is bounded by $c_0$ and $c_{t-1}$. 

Following, to avoid over-fitting on source $n$-gram vectors, we add noise to them. This brings variability to the training process to let the network adapt to unseen target $n$-gram vectors later at inference. Let $\bm{\epsilon}_t$ be a white noise:
\begin{equation}
    \bm{\epsilon}_t = [\epsilon_{t,1}, \ldots, \epsilon_{t,K} ]^{\top}, \, \epsilon_{t,k} \sim \mathcal{U}(a,\,b) \\
\end{equation}
This noise is applied at step $t$ with sample-wise probability $\tau$:
\begin{equation}
    \phi(\bm{s}^\text{NGI}_t) = 
        \begin{cases}
        \psi(\bm{s}^\text{NGI}_t + \bm{\epsilon}_t), & \text{if}\ X < \tau, \, X \sim \mathcal{U}(0,\,1) \\
        \bm{s}^\text{NGI}_t, & \text{otherwise} 
        \end{cases} 
\label{eq:noise}
\end{equation}
where $\phi : \mathbb{R}^{K} \longrightarrow \mathbb{R}^{K}$ is the function applying noise with probability $\tau$, and $\psi : \mathbb{R}^{K} \longrightarrow \mathbb{R}^{K}$ normalizes back to a valid probability distribution after alteration by the noise $\bm{\epsilon}_t$.

Finally, at step $t$, $n$-gram vectors resulting from Equation~\ref{eq:noise} from all previous steps are concatenated into a matrix $\bm{S}^\text{NGI} = [\bm{s}^\text{NGI}_1, \ldots, \bm{s}^\text{NGI}_t]^{\top}$. This matrix $\bm{S}^\text{NGI}$ is given as input to a feed-forward projection then summed with the embeddings of output characters from previous steps $\xi(\bm{c})$ and positional encoding $\bm{P}$. We formalize the injection into transformer input $\bm{X}$ at decoding step $t$ as:
\begin{equation}
\bm{X} = f(\phi(\bm{S}^\text{NGI})) + \xi(\bm{c}) + \bm{P}
\end{equation}
where $\phi : \mathbb{R}^{t \times K} \longrightarrow \mathbb{R}^{t \times K}$ is the same noise function as before but in the matrix domain, $f : \mathbb{R}^{t \times K} \longrightarrow \mathbb{R}^{t \times d}$ is a feed-forward projection, $\bm{c}= [c_0, \ldots, c_{t-1}]^{\top}$ is the concatenation of previous characters, $\xi : \mathbb{R}^{t} \longrightarrow \mathbb{R}^{t \times d}$ and $\bm{P} \in \mathbb{R}^{t \times d}$ refer to an embedding function and positional encoding as in the original transformer~\cite{vaswaniAttentionAllYou2017}, respectively, and $d$ is the internal dimension of the transformer decoder. 

\subsubsection{Discussion}

We choose $n$-gram models as external LM because they exhibit good properties for providing a neural architecture with an LM at a low computational cost. They are indeed lightweight and condensed representations of the language from a given corpus. Also, they can be estimated quickly as $n$-gram counting can be done in a couple minutes on a standard CPU. As opposed to sampling one discrete character (e.g. the one with maximum probability) from the $n$-gram, we argue that preserving the empirical distribution across all characters provides the NN with richer information. Besides, this framework is independent from the order $n$ of the $n$-gram, meaning that any $n$ can be used. 

Injecting the LM into the input of the auto-regressive transformer decoder brings several benefits. 

First, if a network models language on source data as is the case for standard transformers in the literature, then it keeps obeying the source language and make errors once evaluated on a different target corpus. Teaching the network to make use of a provided external LM instead allows to switch it seamlessly at inference time, without any modification to the network's parameters---not even of the projection $f$. This also means that it does not require any extra training on target image-text pairs, meaning we can leverage unpaired text data. By keeping the standard auto-regressive re-injection of previously predicted characters into the transformer decoder along with NGI, we allow the network to still learn some share of biased language rules during training, e.g. the English or French syntax, which remain relevant on the target subset. 

Second, by injecting the LM at the very beginning of the decoder, we allow the transformer to fully integrate and benefit from the language information. Importantly, there is no hyper-parameter to manually tune since the network learns to control the appropriate importance to give to the $n$-gram LM, balancing it dynamically with encoder features to resolve visual ambiguities. By moving away from post-processing, we avoid re-scoring computations which otherwise induce a sensible time complexity overhead during inference for they explore a large search graph or lattice. Moreover, post-processing does not let the NN adapt to the $n$-gram during training. 

Third, setup and inference are low-cost. The overhead caused by NGI is minimal since $n$-gram vectors have a low inference cost and can even be stored in cache to improve memory reads. The parameters and computing cost of the additional feed-forward projection $f$ is also negligible compared to the rest of the network which we leave as is. 

\subsection{Word attention network}

We now introduce the word attention network (WAN), an FCN-transformer built as a smaller counterpart of the well-known, full-page HTR model DAN~\cite{coquenetDANSegmentationfreeDocument2023}. With 2.1M parameters, the WAN is ten times smaller. We note that WAN is merely one example of an architecture into which NGI can be plugged. A lightweight architecture is more relevant in our setting as our training corpus is not sufficiently large to train bigger models. Also, the handwritten word recognition task is easier than full page HTR. We choose an FCN encoder because FCN are efficient for image feature extraction and can process variable input shapes, which is suited for scanned text images. For the decoder part, we opt for a transformer since such attention-based models are commonly used in state-of-the-art approaches in HTR. Full architectural details are given in Section~\ref{sec:training_details}.

\section{Datasets}

As shown in Table~\ref{table:datasets}, we conduct experiments on a total of five handwritten word-level dataset splits, four of which are custom splits of the well-known IAM~\cite{martiIAMdatabaseEnglishSentence2002} and RIMES~\cite{grosickiICDAR2011French2011} datasets while the latter is an industrial use case on name and surname recognition in forms. The custom splits are provided to exacerbate a language shift and its associated bias, as discussed later on. Additionally, we report the $5$-gram character-level perplexity (PPL) on test sets to assess the strength of the shift between source and target language distributions, where a higher PPL denotes a bigger shift. Third to last and second to last columns use a $5$-gram estimated on the source development set, i.e. the union of the training and validation sets. Last column uses a $5$-gram estimated on the target $n$-gram set, denoted by NG. This target NG set was created to estimate our $n$-grams, ensuring that they are not computed directly on target test sets.

\subsection{Perplexity-controlled splits on IAM \& RIMES}

The IAM~\cite{martiIAMdatabaseEnglishSentence2002} dataset is a collection of modern documents in English. Word text-image pairs are from the Aachen split, also known as IAM-A.
Regarding the RIMES dataset, we use the 2011 version~\cite{grosickiICDAR2011French2011} which consists of French administrative letters. Word images are extracted from the body of the letters, excluding addresses and signatures. We provide custom splits of IAM and RIMES as official splits do not naturally exhibit a language shift. We consider two splitting strategies, referred to as \textit{lexicon} and \textit{$k$-means} splits. As reported in Table~\ref{table:datasets}, source training, validation and test set respectively contain 70\%, 7.5\% and 7.5\% of the samples. The remaining 15\% percent are equally shared between the target NG and test sets. All splits are publicly released\footnote{\url{https://www.kaggle.com/datasets/papericdar2026/iam-and-rimes2011-custom-splits}.}.

\begin{table}[t]
\caption{\textbf{Size and perplexity of the five word-level HTR dataset splits.} Dev. denotes the source development set, NG the target $n$-gram set.}
\label{table:datasets}
\centering
\setlength{\tabcolsep}{2pt}
\begin{tabular}{ccccccccc}
\toprule
\multirow{3}[3]{*}{\textbf{Dataset}} & \multirow{3}[3]{*}{\textbf{Split}} & \multicolumn{4}{c}{\textbf{Size}} & \multicolumn{3}{c}{\textbf{$5$-gram PPL}} \\ 
\cmidrule(lr){3-6} \cmidrule(lr){7-9} 
 & & \multicolumn{2}{c}{\textbf{Source}} & \multicolumn{2}{c}{\textbf{Target}} & \multicolumn{2}{c}{\textbf{Source dev.}} & \textbf{Target NG} \\
\cmidrule(lr){3-4} \cmidrule(lr){5-6} \cmidrule(lr){7-8} \cmidrule(lr){9-9}
 & & \textbf{Dev.} & \textbf{Test} & \textbf{NG} & \textbf{Test} & \textbf{Source} & \textbf{Target} & \textbf{Target} \\
\midrule
\multirow{1.5}[1]{*}{IAM} & Lexicon & 69,595 & 6,507 & 6,511 & 7,187 & 4.6 & 14.5 & 4.3 \\
 & $k$-means & 69,595 & 6,730 & 6,511 & 6,964 & 4.6 & 26.2 & 3.8 \\
\hline \\[-2ex]
\multirow{1.5}[1]{*}{RIMES} & Lexicon & 51,908 & 5,022 & 4,857 & 5,191 & 3.1 & 8.3 & 2.4 \\
 & $k$-means & 51,908 & 5,670 & 4,857 & 4,543 & 3.1 & 66.4 & 2.4 \\
\hline \\[-2ex]
N2S (private) & --- & 119,735 & 9,101 & 879,350 & 9,635 & 3.5 & 24.5 & 18.9 \\
\bottomrule 
\end{tabular}
\end{table}

\subsubsection{Lexicon split}

A way to ensure a significant shift between source and target languages is to divide the lexicon $\mathcal{L}$ of the whole dataset into two, that is one lexicon for the source subset $\mathcal{L}_S$ and another for the target subset $\mathcal{L}_T$, such that $\mathcal{L}_S \cap \mathcal{L}_T = \varnothing$. 

\subsubsection{$k$-means split}

Another means of creating a shift between languages is to embed words into a latent space in which we separate the projections into two clusters. We represent each word $w$ of the lexicon $\mathcal{L}$ by a boolean vector $\bm{v}_w$ where $v_{w,i}$=1 if the 2-gram $i$ is present in $w$, so that the size of $\bm{v}_w$ equals the number of 2-grams in the corpus. Then, we split those vectors into two clusters using the $k$-means algorithm with $k=2$. Here, $\mathcal{L}_S \cap \mathcal{L}_T \neq \varnothing$, yet the source-development-to-target-test PPL is higher than that of the lexicon split. It is expected that the source-to-target performance gap will be larger on the $k$-means split than on the lexicon split. Indeed, we assume a higher source-to-target PPL expresses a more pronounced language bias and thus presumes a more challenging recognition. 

\subsection{N2S private dataset}

\begin{figure}
    \centering
    \includegraphics[width=0.75\linewidth]{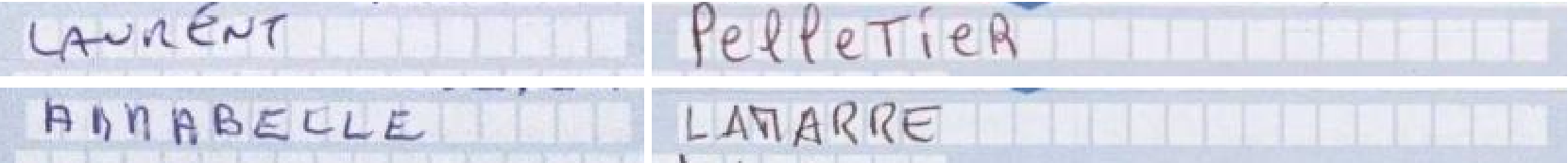}
    \caption{Scans of names (left) and surnames (right) from the N2S dataset.}
    \label{fig:name2surname}
\end{figure}

The name to surname private dataset, denoted by N2S, is a real-world scenario of name and surname recognition in handwritten complaint forms, four samples of which are shown in Fig.~\ref{fig:name2surname}. In this industrial dataset, the name and surname of complaining users must be read. For this purpose, NG is an external database containing a census of surnames with their number of occurrences. 

\section{\label{expe}Experiments}

\subsection{Evaluation protocol}

In our experiments, we compare the proposed WAN to three encoder-decoders to estimate the state-of-the-art transfer performance of biased architectures: TrOCR~\cite{liTrOCRTransformerbasedOptical2023}, a well-known transformer-only architecture for text line recognition; the DAN~\cite{coquenetDANSegmentationfreeDocument2023}, a full-page FCN-transformer HTR model; and SaLT~\cite{meyerRelaxedSyntaxModeling2026}, a debiased and lightweight FCN-transformer designed to recognize arbitrary characters sequences as on vehicle license plates. All three rely on auto-regressive character-level decoders. We report the standard character error rate (CER) metric to evaluate system performance, expressed as a percentage. Regarding NGI LM switching, we use an $n$-gram estimated on the source development set when evaluating on the source test set, whereas evaluation on the target test set uses an $n$-gram estimated on the target $n$-gram set NG (different from the target test set).

\subsection{\label{sec:training_details}Training details}

All models are trained until convergence for 100 epochs on the source training subset only. They are all trained from scratch except TrOCR whose encoder uses pre-trained weights from the first-stage checkpoint\footnote{\url{https://huggingface.co/microsoft/trocr-small-stage1}.} since it could not achieve satisfying results on such small datasets otherwise. Notably, transformer decoders benefit from training from scratch because pre-training would only bias them towards the source training corpus~\cite{sriramCOLDFUSIONTRAINING2018}. Images keep their original shapes and follow ImageNet normalization for all models except TrOCR which keeps its custom image resizing and pre-processing strategy. Heavy image augmentation is performed in the same way as Coquenet \textit{et al.}~\cite{coquenetDANSegmentationfreeDocument2023} with DPI scaling, perspective change, elastic distortion, erosion, color jittering, Gaussian blur and noise, sharpening and random zooming. There is a 0.9 probability that these augmentations are applied, with a 0.2 probability for each of them. No synthetic data is used. Optimization is done on the cross-entropy loss with the Adam optimizer~\cite{kingmaAdamMethodStochastic2017}. Model selection is performed according to validation CER on the source subset.

We use TrOCR\textsubscript{SMALL} variant throughout this paper. For a fairer comparison, we replace TrOCR's byte-pair-encoding (BPE)~\cite{sennrichNeuralMachineTranslation2016} tokenizer with a character-level one. We then fine-tune the whole TrOCR, with a batch size of 256 and a learning rate of ${2 \times 10^{-5}}$. DAN and SaLT are trained with a batch size of 128 and a learning rate of ${1 \times 10^{-4}}$ and ${3 \times 10^{-4}}$, respectively.

WAN consists in a ten-layer FCN encoder and a small two-layer transformer decoder, down from the $6$ and $8$ decoder layers of TrOCR and the DAN. Each of the ten layers comprises a convolution, ReLU activation, instance normalization~\cite{ulyanovImprovedTextureNetworks2017}, $2 \times 2$ max pooling and dropout \cite{srivastavaDropoutSimpleWay2014} with a probability of 0.1. Each layer's convolution has a $3 \times 3$ kernel, a stride of $1$ and padding in such a way that the output has the same shape as the input. Output channels are increased so that the first three layers output $32$ channels, the next three $64$, the following three $128$ and the final layer $256$. The transformer decoder has an internal dimension $d = 256$ with $8$ heads. We employ teacher forcing error (TFE) with probability $0.1$ in the WAN's transformer decoder during training to simulate wrong previous predictions which might occur at inference time. This demonstrated better generalization abilities in HTR~\cite{coquenetDANSegmentationfreeDocument2023}. TFE also impacts NGI, as changing the context generates alternative $n$-gram vectors. Through simulating erroneous predictions, TFE restrains the network's reliance on both internal language modeling and external LM by reducing its confidence into previous predictions. The WAN is trained with a batch size of 128 and a learning rate of ${3 \times 10^{-4}}$.

Regarding NGI, all $n$-grams referred to in this paper are character-level and use interpolation with Witten-Bell smoothing~\cite{wittenZerofrequencyProblemEstimating1991}. We use a value of $n=5$ unless stated otherwise, as it is a good trade-off between estimation quality and compute time. The SRILM toolkit is used to estimate $n$-grams. White noise $\epsilon$ uses $a=-0.1, b=0.1, \tau=0.2$ following hyper-parameter search on the source subset. The projection $f$ is a feed-forward layer followed by layer normalization~\cite{baLayerNormalization2016a} and ReLU activation. 

\subsection{Performance drop of biased state-of-the-art networks}

Tables~\ref{table:sota_iam_rimes} and~\ref{table:sota_name2surname} present results on source and target subsets without NGI. All three state-of-the-art models and the WAN suffer from a significant performance gap between source and target subsets of IAM, RIMES and N2S. When comparing splits in each dataset, we observe that source CER are almost equal while CER approximately double from lexicon to $k$-means target splits, except for SaLT whose debiasing reduces the gap as discussed later on. This doubling in CER is linked to the surge in PPL from source to target that is higher among $k$-means splits than lexicon splits, as reported in Table~\ref{table:datasets}: the higher the difference in PPL between subsets, the larger the gap in CER. In particular, the WAN without NGI suffers from a performance gap similar to state-of-the-art alternatives, rising from 7.6 to 23.4 CER and from 4.3 to 29.9 CER on $k$-means splits of IAM and RIMES, respectively. Likewise for N2S, the CER more than doubles from source to target test set. As expected, the WAN without NGI is no exception with a CER increasing from 2.5 to 6.3. Such poor error rates make a transcription system unusable in practice, which we aim to alleviate thanks to NGI as described in the next section. 

Although TrOCR and the DAN were proposed for line-level and page-level HTR, respectively, we observe that they effortlessly adapt to word-level HTR on source test sets. However, they suffer from a drop comparable to that of the WAN on target test sets, which is expected as they all absorb training corpus bias. Note that TrOCR and the DAN are not evaluated with off-the-shelf pre-trained weights as this would not result in a fair comparison with SaLT and the WAN which must be trained from scratch. The comparatively poor performance of SaLT can be explained by its debiased design which stops syntactical modeling of the training language. This proved to be successful on vehicle license plates which do not follow any rules in the ordering of their characters, however in handwriting recognition, learning language-specific rules is crucial for improving prediction performance.

\begin{table}[t]
\centering
\caption{\textbf{CER(\%) of transformer-based models trained on the four custom dataset splits from IAM and RIMES.} All four models suffer from a significant performance gap between source and target subsets, including the WAN without NGI.}
\label{table:sota_iam_rimes}
\setlength{\tabcolsep}{2pt}
\begin{tabular}{ccccccccc}
\toprule
\multirow{3}[3]{*}{\textbf{Model}} & \multicolumn{4}{c}{\textbf{IAM}} & \multicolumn{4}{c}{\textbf{RIMES}} \\ 
\cmidrule(lr){2-5} \cmidrule(lr){6-9}
 & \multicolumn{2}{c}{\textbf{Lexicon split}} & \multicolumn{2}{c}{\textbf{$k$-means split}} & \multicolumn{2}{c}{\textbf{Lexicon split}} & \multicolumn{2}{c}{\textbf{$k$-means split}} \\
\cmidrule(lr){2-3} \cmidrule(lr){4-5} \cmidrule(lr){6-7} \cmidrule(lr){8-9}
 & \textbf{Source} & \textbf{Target} & \textbf{Source} & \textbf{Target} & \textbf{Source} & \textbf{Target} & \textbf{Source} & \textbf{Target} \\ 
\midrule
TrOCR~\cite{liTrOCRTransformerbasedOptical2023} & 6.3 & 12.9 & 6.7 & 21.4 & 3.3 & 12.9 & 3.6 & 33.3 \\
DAN~\cite{coquenetDANSegmentationfreeDocument2023} & 6.0 & 15.0 & 6.1 & 26.1 & 3.7 & 16.8 & 3.9 & 40.2 \\
SaLT~\cite{meyerRelaxedSyntaxModeling2026} & 15.5 & 25.1 & 15.5 & 29.8 & 10.7 & 25.9 & 12.3 & 38.6 \\
WAN\tablefootnote{\label{note:table_note}For a fair comparison, here the WAN is trained without TFE to match the training strategy of other models.} & 6.9 & 15.8 & 7.6 & 23.4 & 4.7 & 18.1 & 4.3 & 29.9 \\
\bottomrule
\end{tabular}
\end{table}

\begin{table}[t]
\caption{\textbf{CER(\%) of transformer-based models trained on the name to surname (N2S) dataset.} As on IAM and RIMES, all four models suffer from a performance gap between source and target corpora, including the WAN without NGI.}\label{table:sota_name2surname}
\centering
\setlength{\tabcolsep}{4pt}
\begin{tabular}{ccc}
\toprule
\textbf{Model} & \textbf{Source} & \textbf{Target} \\ 
\midrule
TrOCR~\cite{liTrOCRTransformerbasedOptical2023} & 2.2 & 7.1 \\
DAN~\cite{coquenetDANSegmentationfreeDocument2023} & 2.1 & 5.3 \\
SaLT~\cite{meyerRelaxedSyntaxModeling2026} & 4.0 & 13.4 \\
WAN\textsuperscript{\ref{note:table_note}} & 2.5 & 6.3 \\
\bottomrule
\end{tabular}
\end{table}

\subsection{Reducing the performance gap}

We now investigate the positive impact on performance retention provided by NGI into the WAN. Table~\ref{table:main_results_iam_rimes} shows that NGI ensures a lower target CER on all four custom splits of IAM and RIMES while preserving source CER. 
The WAN+NGI performs better on target corpora than the WAN alone and than state-of-the-art models from Table~\ref{table:sota_iam_rimes}. Indeed, the WAN+NGI reaches CER of 11.8 and 10.1 on the lexicon and $k$-means target test sets for IAM, respectively. On RIMES target test sets, it obtains a CER of 8.8 on the lexicon split and 19.2 on the $k$-means split. A similar trend is reported on N2S in Table~\ref{table:main_results_names} where the WAN+NGI obtains a target test CER of 4.5, proving that the proposed injection of a relevant $n$-gram helps reading out-of-distribution handwriting scans. 

\begin{table}[t]
\caption{\textbf{CER(\%) of the proposed NGI compared to the WAN alone on the four custom dataset splits from IAM and RIMES.} NGI consistently reduces the performance gap from source to target subsets. A post-processing LM, denoted by +LM, further reduces the gap. We highlight in bold the best results on target test sets as they are our main objective.}
\label{table:main_results_iam_rimes}
\centering
\setlength{\tabcolsep}{1pt}
\begin{tabular}{ccccccccc}
\toprule
\multirow{3}[3]{*}{\textbf{Model}} & \multicolumn{4}{c}{\textbf{IAM}} & \multicolumn{4}{c}{\textbf{RIMES}} \\ 
\cmidrule(lr){2-5} \cmidrule(lr){6-9}
 & \multicolumn{2}{c}{\textbf{Lexicon split}} & \multicolumn{2}{c}{\textbf{$k$-means split}} & \multicolumn{2}{c}{\textbf{Lexicon split}} & \multicolumn{2}{c}{\textbf{$k$-means split}} \\
\cmidrule(lr){2-3} \cmidrule(lr){4-5} \cmidrule(lr){6-7} \cmidrule(lr){8-9}
 & \textbf{Source} & \textbf{Target} & \textbf{Source} & \textbf{Target} & \textbf{Source} & \textbf{Target} & \textbf{Source} & \textbf{Target} \\ 
\midrule
WAN & 6.0 & 13.2 & 6.8 & 18.0 & 3.4 & 12.2 & 4.0 & 25.4 \\
WAN+NGI & 6.2 & \textbf{11.8} & 6.8 & \textbf{10.1} & 3.5 & \textbf{8.8} & 3.6 & \textbf{19.2} \\
\hline \\[-2ex]
WAN+LM & 5.9 & \textbf{10.6} & 6.8 & 9.7 & 3.3 & 6.6 & 3.7 & 13.8 \\
WAN+NGI+LM & 6.5 & 11.5 & 6.9 & \textbf{8.4} & 3.4 & \textbf{5.1} & 3.7 & \textbf{10.7} \\
\bottomrule
\end{tabular}
\end{table}

\begin{table}[t]
\caption{\textbf{CER(\%) of the proposed NGI compared to the WAN alone on the name to surname (N2S) dataset.} As on IAM and RIMES, NGI reduces the drop on the target subset. We highlight in bold the best result on the target test set as it is our main objective.}
\label{table:main_results_names}
\centering
\setlength{\tabcolsep}{4pt}
\begin{tabular}{ccc}
\toprule
\textbf{Model} & \textbf{Source} & \textbf{Target} \\ 
\midrule
WAN & 2.2 & 5.0 \\
WAN+NGI & 2.4 & \textbf{4.5} \\
\bottomrule
\end{tabular}
\end{table}

Moreover, NGI allows the network to better compensate the higher bias on $k$-means splits. Notably, on IAM, the WAN suffers from a source-to-target gap from 6.8 to 18.0 CER on the more challenging $k$-means split, which is higher than the increase from 6.0 to 13.2 CER on the lexicon split. On the contrary, the WAN+NGI has a smaller gap on the $k$-means split, from 6.8 to only 10.1 CER, than on the lexicon split which shows an increase from 6.2 to 11.8 CER.
Nonetheless, the performance gain on the N2S dataset is minimal which can be justified by a much higher target PPL of 18.9 as displayed in Table~\ref{table:datasets}. A quick analysis of the surnames census reveals that this database contains erroneous fields that do not correspond to surnames, for example numerical values.

An additional post-processing LM, denoted by +LM in Table~\ref{table:main_results_iam_rimes}, further improves target performance on IAM and RIMES. This LM has the same configuration as the one used for NGI. Lattice re-scoring is done using the SRILM toolkit with beam width equal to 150.
Both the WAN and WAN+NGI benefit from applying an LM for post-processing. Note that using the $n$-gram as a post-processing (WAN+LM) is slightly better than injecting into the input of the network (WAN+NGI), at the cost of an increased computational complexity. Finally, the combination of the two $n$-gram methods into a WAN+NGI+LM attains the best performance on the target test set. 

\begin{table}[t]
\caption{\textbf{Ablation study of our proposal (WAN+NGI) by removing specific components of NGI and reducing the $n$-gram order on the $k$-means splits from IAM and RIMES.} We report CER(\%) when removing noise or the TFE training strategy from NGI, and lowering $n$ to $\{3,2\}$ for NGI. Ablations are not cumulative.}
\label{table:ablation}
\centering
\setlength{\tabcolsep}{4pt}
\begin{tabular}{lcccc}
\toprule
\multicolumn{1}{c}{\multirow{2}[2]{*}{\textbf{Model}}} & \multicolumn{2}{c}{\textbf{IAM}} & \multicolumn{2}{c}{\textbf{RIMES}} \\
\cmidrule(lr){2-3} \cmidrule(lr){4-5} 
 & \textbf{Source} & \textbf{Target} & \textbf{Source} & \textbf{Target} \\ \midrule
WAN+NGI (5-gram) & 6.8 & 10.1 & 3.6 & 19.2 \\
\hline \\[-2ex]
- Noise $\phi$ & 6.6 & 10.6 & 3.7 & 19.2 \\
- TFE & 10.3 & 16.5 & 5.9 & 20.8 \\
\hline \\[-2ex]
3-gram & 6.8 & 18.1 & 3.2 & 26.2 \\
2-gram & 6.7 & 19.3 & 3.3 & 25.5 \\
\bottomrule
\end{tabular}
\end{table}

\subsection{Ablation study of NGI components}

In Table~\ref{table:ablation}, we present results after separately ablating NGI components and changing the order $n$ of the injected $n$-gram. We perform the ablation study on the $k$-means splits from IAM and RIMES as their higher perplexity makes them more challenging. Removing noise slightly degrades CER as the network tends to overfit on the training corpus. Notably, $\bm{s}^\text{NGI}_1$ is computed from the start-of-sequence token as context, producing a constant vector $\bm{s}^\text{NGI}_1$ for a given LM. Thus, adding noise to $\bm{s}^\text{NGI}_1$ provides variability when learning to predict the first token, improving adaptability to new LM during inference. Removing TFE reduces generalization capabilities regarding both auto-regressively injected tokens and $n$-grams, altering CER on both target and source test sets. Switching to lower-order 3-grams or 2-grams causes the largest performance drop on target corpora. It is expected that the closer the $n$-gram estimate of the underlying target distribution, the smaller the performance drop. In these cases, network training is done with an NGI of the same order $n$ as that injected at test time. 

\section{Conclusion \& future work}

This paper investigates the robustness of transformer-based encoder-decoders to language shift in word-level HTR. After revealing the recognition performance drop of state-of-the-art architectures when the test language distribution differs from training, we propose n-gram injection (NGI) as a method for dynamically injecting an external $n$-gram in order to reduce the error surge. In addition, we introduce the word attention network (WAN) as a lightweight FCN-transformer for handwritten word recognition, in which NGI is applied. Through experiments, NGI demonstrates successful character recognition under linguistic distribution shifts, complementing the transformer decoder of the WAN. By switching to an appropriate $n$-gram at test time, our approach mitigates bias without any extra training of the neural recognizer. Comparing results with lattice re-scoring suggests that NGI is a promising step towards moving away from costly post-processing. Lastly, we expect our framework to be compatible with neural external language models, which future work could for instance leverage for challenging cross-lingual transfer if computational complexity is less of an issue.

\begin{credits}
\subsubsection{\ackname} This study was funded by the CIFRE ANRT grant No. 2023/0195 and contributes to AI Excellence Cluster SequoIA (grant ANR-23-IACL-0009).
\end{credits}

\bibliographystyle{splncs04}
\bibliography{biblio}

\end{document}